\documentclass[sigconf]{acmart}

\usepackage{subfigure}
\usepackage{algorithm}
\usepackage{algorithmic}
\usepackage{multirow}
\usepackage{float}

\AtBeginDocument{%
  }

\setcopyright{acmcopyright}
\copyrightyear{2023}
\acmYear{2023}
\acmDOI{XXXXXXX.XXXXXXX}

\acmConference[WWW'23]{the Web Conference}{April 30-May 4,2023}{Austin, Texas, USA}
\acmPrice{15.00}
\acmISBN{978-1-4503-XXXX-X/18/06}

\begin{document}

\title{INCREASE: Inductive Graph Representation Learning for Spatio-Temporal Kriging}


\author{Chuanpan Zheng}
\affiliation{%
  \institution{Fujian Key Laboratory of Sensing and Computing for Smart Cities, School of Informatics, Xiamen University}
  \city{}
  \country{}
}
\email{zhengchuanpan@stu.xmu.edu.cn}

\author{Xiaoliang Fan}
\affiliation{%
  \institution{Fujian Key Laboratory of Sensing and Computing for Smart Cities, School of Informatics, Xiamen University}
  \city{}
  \country{}
}
\email{fanxiaoliang@xmu.edu.cn}

\author{Cheng Wang}
\authornote{Corresponding author.}
\affiliation{%
  \institution{Fujian Key Laboratory of Sensing and Computing for Smart Cities, School of Informatics, Xiamen University}
  \city{}
  \country{}
}
\email{cwang@xmu.edu.cn}

\author{Jianzhong Qi}
\affiliation{%
  \institution{School of Computing and Information Systems, University of Melbourne}
  \city{}
  \country{}}
\email{jianzhong.qi@unimelb.edu.au}

\author{Chaochao Chen}
\affiliation{%
 \institution{College of Computer Science and Technology, Zhejiang University}
 \city{}
 \country{}}
\email{zjuccc@zju.edu.cn}

\author{Longbiao Chen}
\affiliation{%
 \institution{Fujian Key Laboratory of Sensing and Computing for Smart Cities, School of Informatics, Xiamen University}
 \city{}
 \country{}}
\email{longbiaochen@xmu.edu.cn}

\begin{abstract}
  Spatio-temporal kriging is an important problem in web and social applications, such as Web or Internet of Things, where things (e.g., sensors) connected into a web often come with spatial and temporal properties. It aims to infer knowledge for (the things at) unobserved locations using the data from (the things at) observed locations during a given time period of interest. This problem essentially requires \emph{inductive learning}. Once trained, the model should be able to perform kriging for different locations including newly given ones, without retraining. However, it is challenging to perform accurate kriging results because of the heterogeneous spatial relations and diverse temporal patterns. In this paper, we propose a novel inductive graph representation learning model for spatio-temporal kriging. We first encode heterogeneous spatial relations between the unobserved and observed locations by their spatial proximity, functional similarity, and transition probability. Based on each relation, we accurately aggregate the information of most correlated observed locations to produce inductive representations for the unobserved locations, by jointly modeling their similarities and differences. Then, we design relation-aware gated recurrent unit (GRU) networks to adaptively capture the temporal correlations in the generated sequence representations for each relation. Finally, we propose a multi-relation attention mechanism to dynamically fuse the complex spatio-temporal information at different time steps from multiple relations to compute the kriging output. Experimental results on three real-world datasets show that our proposed model outperforms state-of-the-art methods consistently, and the advantage is more significant when there are fewer observed locations. Our code is available at https://github.com/zhengchuanpan/INCREASE.
\end{abstract}



\maketitle
\section{Introduction}

With recent advances in the Web of Things (WoT)~\cite{Hui-et-al:WWW2022,Xu-et-al:TMC2022,Chen-and-Weng:TMC2022}, more and more devices are collecting various types of location-based data in cities, e.g., check-in data in location-based social networks (LBSN)~\cite{Zhang-et-al:WWW2022,Schweimer-et-al:WWW2022}, Internet of Things (IoT) sensor data~\cite{Zhang-et-al:TITS2022,Liang-et-al:IJCAI2018}. Due to the high operating cost, the number of available devices is often still limited and the device distribution is usually unbalanced. This makes it difficult to offer fine-grained and high spatial-resolution analysis using the observation data only. \emph{Spatio-temporal kriging} aims to address this data sparsity and skewed data availability problem -- using the data from observed locations to mine the patterns or infer the knowledge for unobserved locations during a time period of interest~\cite{Wu-et-al:AAAI2021}. Despite recent developments in spatio-temporal data mining~\cite{Wang-et-al:TKDE2022,Zheng-et-al:TITS2022,Wu-et-al:IJCAI2022}, little attention has been paid to the spatio-temporal kriging, which is an important problem in web applications (see Appendix~\ref{appendix relevance to web}). We expect a solution for this problem to generate a broad impact over web-based technologies in transport industry~\cite{Deng-et-al:TITS2022}, urban planning~\cite{Wu-et-al:IJCAI2022}, and socio-economic system~\cite{Schweimer-et-al:WWW2022}, as well as the web data mining~\cite{He-et-al:WWW2022}.


There are two types of methods for spatio-temporal kriging: transductive models and inductive ones. The transductive models, such as matrix/tensor completion approaches~\cite{Bahadori-et-al:NeurIPS2014,Deng-et-al:TITS2022}) require retraining for new graph structures.
Inductive models, such as \emph{graph convolutional networks} (GCN)~\cite{Wu-et-al:TNNLS2021} based approaches~\cite{Appleby-et-al:AAAI2020,Wu-et-al:AAAI2021} are able to accommodate dynamic graph structures. Once trained, they can perform kriging for different unobserved locations including newly given ones without retraining. However, there still lacks a satisfactory progress in the spatio-temporal kriging problem, mainly due to the following challenges.

\begin{figure*}
	\centering
	\subfigure[An example of heterogeneous spatial relations.]{
		\label{fig1(a)} 
		\includegraphics[width = 0.4 \textwidth]{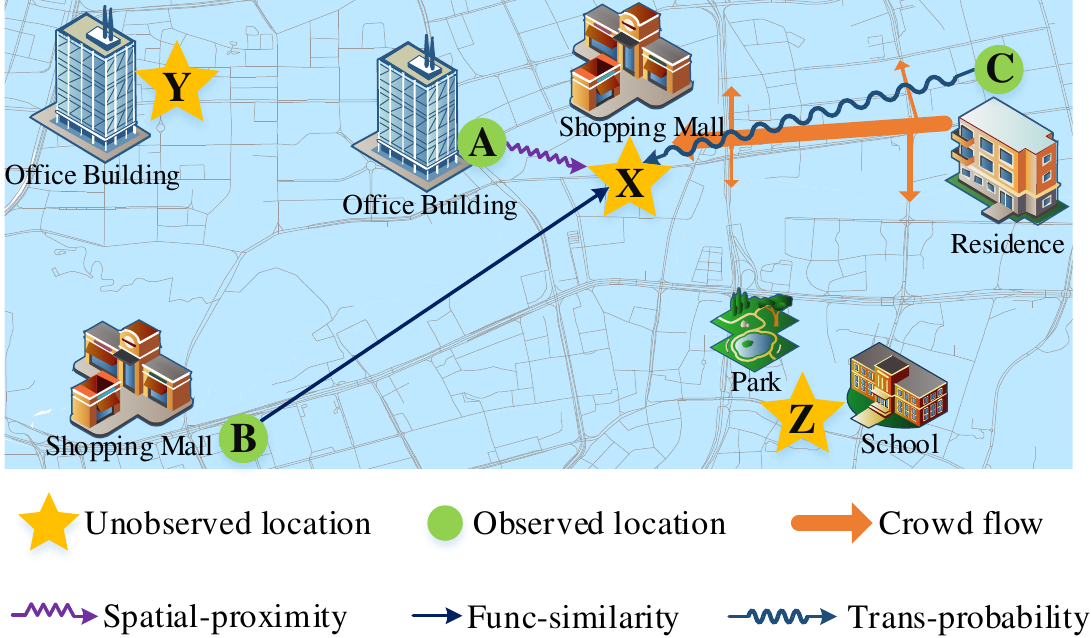}}		
	\subfigure[An example of diverse temporal patterns.]{
		\label{fig1(b)} 
		\includegraphics[width = 0.42 \textwidth]{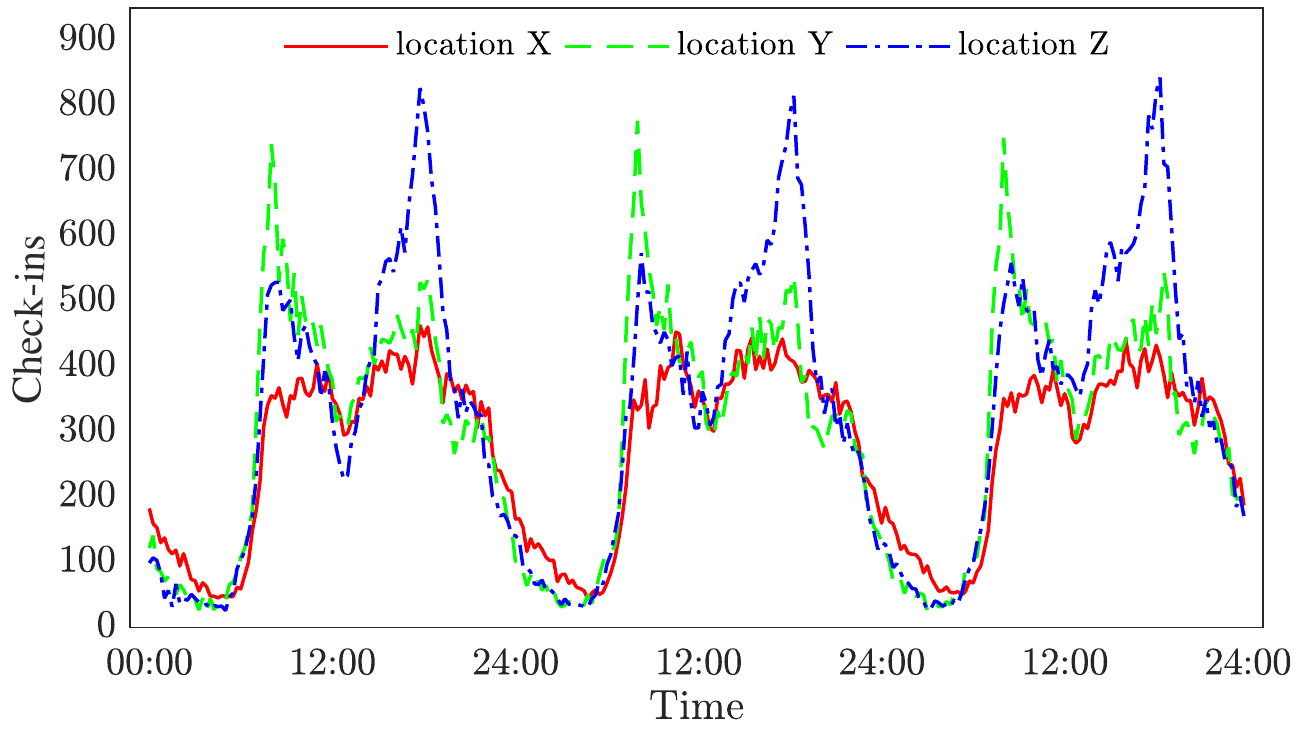}}
	\caption{A motivating example. (a) Heterogeneous spatial relations: spatial proximity relation (e.g., people work at office location $ A $ may shop at a near mall $ X $), functional similarity relation (e.g., locations $ B $ and $ X $ are both close to shopping malls, and they may share similar check-in patterns), and transition probability relation (e.g., a large crowd of people may transit from a residential area $ C $ to shopping mall $ X $ due to the convenient public transport). (b) Diverse temporal patterns: the check-in pattern at location $ X $ is relatively steady in the day time, while locations $ Y $ and $ Z $ have morning and evening peaks, respectively.}
	\label{fig1}
\end{figure*}

\textit{Challenge 1}: How to accurately select most correlated locations for each unobserved location? There exists complex relations among different locations in LBSN. For example, in Figure~\ref{fig1(a)}, the unobserved location $ X $ may be correlated to observed locations $ A $, $ B $, and $ C $ because of spatial proximity (e.g., people work at office location $ A $ may shop at a near mall $ X $), similar functionality (e.g., locations $ B $ and $ X $ are both close to shopping malls)
, and transition probability (e.g., a large crowd of people may transit from a residential area $ C $ to shopping mall $ X $ due to the convenient public transport), respectively. Moreover, \textit{two related locations may not be highly correlated all the time}. For example, the check-in patterns at locations $ A $ and $ X $ may be similar at office hours but different at other hours.

\textit{Our solution: Aggregating heterogeneous spatial relations by jointly modeling similarities and differences}. We first encode heterogeneous relations between unobserved and observed locations based on spatial proximity, functional similarity, and transition probability relations. Then, in each relation, we model both similarities and differences between unobserved and observed locations, to aggregate related locations' information to form inductive representations for the unobserved location. 

\textit{Challenge 2}: How to adaptively extract the temporal patterns for each unobserved location? Due to the lack of historical data, it is non-trivial to directly model the temporal correlations for unobserved locations. Moreover, \textit{different unobserved locations may have diverse temporal patterns, which makes this problem more challenging}. For example, in Figure~\ref{fig1(b)}, the check-in pattern at location $ X $ is relatively steady in the day time, while locations $ Y $ and $ Z $ have morning and evening peaks, respectively.

\textit{Our solution: Modeling relational temporal correlations by introducing relation-aware gated recurrent unit (GRU) networks}. We first exploit a relation-aware input gate to adaptively control the importance of the current representation, and a relation-aware forget gate to adaptively control the influence of past states. Then, we assemble two relation-aware gates into GRU networks to guide the temporal information flow for the unobserved location.

\textit{Challenge 3}: How to dynamically combine heterogeneous spatial relations with diverse temporal patterns? The importance of different relations may change quickly. For example, in Figure~\ref{fig1(a)}, to estimate the check-ins of the unobserved location $ X $, the spatial proximity relation (location $ A $) may dominant during office hours, while the transition probability relation (location $ C $) may be more important after office hours.

\textit{Our solution: Fusing complex spatio-temporal information by designing a multi-relation attention mechanism}. We first compute the attention scores of multiple relations at different time steps. Then, we dynamically assign different relations with different weights to fuse the multi-relation information, and compute the final kriging sequence for the unobserved location.

To address these challenges, overall, we propose an \emph{inductive graph representation learning model for spatio-temporal kriging} (INCREASE), based on the aforementioned solutions. INCREASE consists of three stages: spatial aggregation to address Challenge 1, temporal modeling to address Challenge 2, and multi-relation fusion to address Challenge 3. Experimental results on three real-world spatio-temporal datasets demonstrate that our model achieves the best performances against five state-of-the-art competitors, and the advantage is more significant when fewer observed locations are available. The contributions of this paper are summarized as follows.

\begin{itemize}
	\item We propose an inductive graph representation learning model named INCREASE that models heterogeneous spatial relations and diverse temporal patterns for spatio-temporal kriging problems.
	\item We design a multi-relation attention mechanism that can dynamically fuse complex spatio-temporal information at different time steps from different relations to compute the final kriging sequence.
	\item We evaluate the performance of our model on three real-world datasets. The experimental results show that our model outperforms state-of-the-art competitors, especially 14.0\% on MAPE, and the advantage is more significant when there are fewer observed locations are available. 
\end{itemize}



\section{Related Work} \label{Related Work}

Spatio-temporal kriging falls into a broader area of spatio-temporal data mining~\cite{Wang-et-al:TKDE2022}. Recent studies using deep learning have shown promising results in many spatio-temporal data mining problems, such as prediction~\cite{Li-et-al:AAAI2021,Mallah-et-al:TMC2022}, anomaly detection~\cite{Wang-et-al:ICWS2017,Chen-et-al:CBD2018}, recommendation~\cite{Chang-et-al:IJCAI2018,Wang-et-al:KDD2019}, and kriging~\cite{Appleby-et-al:AAAI2020,Wu-et-al:AAAI2021}. We review studies on two most relevant problems -- spatio-temporal prediction~\cite{Zheng-et-al:arXiv2021} and kriging~\cite{Wu-et-al:AAAI2021}. 

\subsection{Spatio-Temporal Prediction} 

Spatio-temporal prediction aims to predict future status (e.g., sensor readings) of a given set of locations based on their historical observations~\cite{Geng-et-al:AAAI2019,Wang-et-al:WWW2020,Zhou-et-al:AAAI2021}. 
\emph{Recurrent neural networks} (RNN), e.g., \emph{long short-term memory} (LSTM) and GRU networks, are commonly used in spatio-temporal prediction tasks for modeling the temporal dependencies~\cite{Sutskever-et-al:NeurIPS2014,Liu-et-al:ICDE2021}. To model the spatial dependencies, \emph{convolutional neural networks} (CNN) are often used~\cite{Zhang-et-al:AAAI2017,Yao-et-al:AAAI2018,Zheng-et-al:TITS2020,Jin-et-al:ICDE2022}. Recent studies formulate spatio-temporal prediction as a graph prediction problem~\cite{Li-et-al:ICLR2018,Yu-et-al:IJCAI2018,Zheng-et-al:AAAI2020}. By exploiting the strong learning capability of graph neural networks~\cite{Wu-et-al:TNNLS2021,Dong-et-al:WWW2022}, these graph based methods have produced promising results~\cite{Chu-et-al:TMC2022,He-et-al:TMC2022,He-et-al:WWW2022,He-and-Shin:WWW2022}. Recently, attention~\cite{Vaswani-et-al:NeurIPS2017,Fan-et-al:WWW2022} based models have shown superior performance in spatio-temporal prediction problems~\cite{Zheng-et-al:AAAI2020,Cirstea-et-al:ICDE2022}. A few other studies learn multiple correlations to improve the prediction accuracy~\cite{Geng-et-al:AAAI2019,Zhou-et-al:AAAI2021}.

While spatio-temporal prediction shares similarity with spatio-temporal kriging in modeling the spatio-temporal correlations, it cannot be directly applied to the spatio-temporal kriging due to no historical data for the unobserved locations. 

\subsection{Spatio-Temporal Kriging}

Compared to the spatio-temporal prediction, less attention has been paid to spatio-temporal kriging. Traditional kriging methods have a strong Gaussian assumption~\cite{Cressie-and-Wikle:2015}, which is quite restrictive, as the data may not follow Gaussian distributions. 

Several studies treat spatio-temporal kriging as a matrix (or tensor) completion problem where the rows corresponding to unobserved locations are completely missing~\cite{Bahadori-et-al:NeurIPS2014,Deng-et-al:TITS2022}. Low-rank tensor models are developed to capture the dependencies among variables~\cite{Bahadori-et-al:NeurIPS2014} or to incorporate spatial autoregressive dynamics~\cite{Deng-et-al:TITS2022}. However, matrix completion is \emph{transductive} which cannot cope with additional locations of interest without model retraining. Another study~\cite{Xu-et-al:TRC2020} uses graph embeddings~\cite{Yan-et-al:TPAMI2007} to select the most relevant observed locations for given unobserved locations and uses \emph{generative adversarial networks} (GAN)~\cite{Goodfellow-et-al:NeurIPS2014} to generate estimations for the unobserved locations. Since graph embedding is transductive, this method is also transductive.

Recent studies~\cite{Appleby-et-al:AAAI2020,Wu-et-al:AAAI2021} make use of the \emph{inductive} power of \emph{graph convolutional networks} (GCN) for inductive kriging. They construct a graph according to the spatial distance between the (unobserved and observed) locations and apply GCN on that graph to recover the values of unobserved locations (nodes on the graph). However, using the distance only may miss the correlations among distant locations.
In this work, we consider heterogeneous spatial relations among the locations, as well as diverse temporal patterns at the unobserved locations, to perform inductive spatio-temporal kriging. 

\section{Spatio-Temporal Kriging} \label{Methodology}

\begin{figure*}
	\centering
	\includegraphics[width=0.9\textwidth]{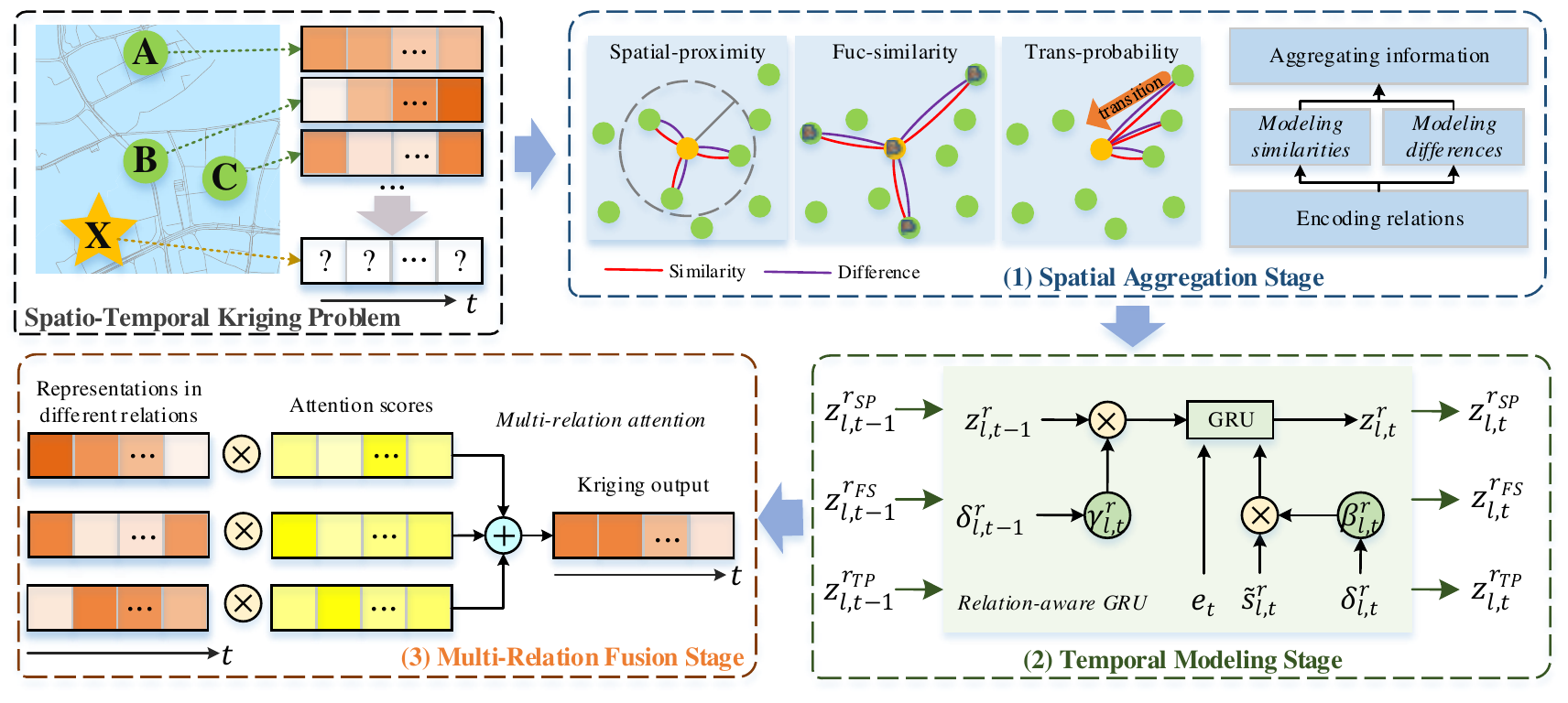} 
	\caption{Overall architecture of INCREASE. A typical spatio-temporal kriging task (the up left part) contains three stages: (1) \textit{spatial aggregation stage}  aggregates related locations' information to form the hidden representation of a target unobserved location $ l $, by jointly modeling their similarities and differences; (2) \textit{temporal modeling stage} adaptively models temporal correlations by introducing a relation-aware GRU network for each relation; and (3) \textit{multi-relation fusion stage} dynamically fuses the representations at different time steps from heterogeneous relations to compute the final kriging sequence.}
	\label{fig2}
\end{figure*}

\subsection{Problem Definition}

Suppose that there are $ N $ observed locations. We use $ \mathbf{x}_{i,t} \in \mathbb{R}^{C} $ to represent $ C $ types of data recorded at time step $ t $ for the $ i $-th location. Given a time window of $ P $ time steps, the data at the $ i $-th location form a (multivariate) time series denoted as 
$ \mathbf{X}_i = (\mathbf{x}_{i,1}, \mathbf{x}_{i,2}, \cdots, \mathbf{x}_{i,P})^\top \in \mathbb{R}^{P \times C} $, while the data of all $ N $ observed locations are denoted as $ \mathcal{X} = (\mathbf{X}_1, \mathbf{X}_2, \cdots, \mathbf{X}_N) \in \mathbb{R}^{N \times P \times C} $.

Given $ \mathcal{X} $, spatio-temporal kriging aims to estimate the time series of $ P $ time steps at $ M $ unobserved locations. We denote the set of estimated time series as $ \mathcal{Y}=(\hat{\mathbf{Y}}_1, \hat{\mathbf{Y}}_2, \cdots, \hat{\mathbf{Y}}_M) \in \mathbb{R}^{M \times P \times C} $, where $ \hat{\mathbf{Y}}_l = {(\hat{\mathbf{y}}_{l,1}, \hat{\mathbf{y}}_{l,2}, \cdots, \hat{\mathbf{y}}_{l,P})}^\top \in \mathbb{R}^{P \times C} $ is the estimated time series for the unobserved location $ l $ ($l=1,\cdots,M$). We provide a table of important notations in Appendix~\ref{appendix notations}.

\subsection{Overview}

Figure~\ref{fig2} shows the structure of our proposed \emph{inductive graph representation learning model for spatio-temporal kriging} (INCREASE). It consists of three stages: (1) \textit{spatial aggregation}, which accurately aggregates the information from correlated locations according to the spatial proximity, functional similarity, and transition probability relations; (2) \textit{temporal modeling}, which adaptively models the temporal dependencies in the generated sequence representations for each relation; (3) \textit{multi-relation fusion}, which dynamically fuses the estimations from different relations to compute the final kriging results. We detail the three stages next. 

\subsection{Spatial Aggregation}

A key problem in spatio-temporal kriging is to identify the observed locations that are closely related to an unobserved location of interest. Most existing studies~\cite{Appleby-et-al:AAAI2020,Wu-et-al:AAAI2021} only use the spatial distance as an indicator of location relevance, which may miss distant yet highly correlated locations. Different from such studies, we leverage multiple heterogeneous relations to infer location relevance.  

\subsubsection{Encoding heterogeneous spatial relations}

We encode three types of relations between unobserved and observed locations.

(1) \textit{Spatial-Proximity.} 
As the spatially closed locations may share similar data patterns~\cite{Appleby-et-al:AAAI2020,Wu-et-al:AAAI2021}, we first use spatial distance to compute the spatial proximity as:
\begin{equation}
\alpha_{i, l}^{r_{SP}} = \exp(-\frac{{dist(i, l)}^2}{\epsilon^2}).
\label{eq1}
\end{equation}
Here, $ dist(i, l) $ denotes the distance from observed location $ i $ to unobserved location $ l $, which can be Euclidean or non-Euclidean (e.g., road network) distance based on application requirements, and $ \epsilon $ is the standard deviation of distances among observed locations.

(2) \textit{Func-Similarity.}
Next, we consider locations correlated by similar functionality. 
For example, the check-in patterns of two shopping malls in different regions may be quite similar. 
To model the functionality of a location, we use a vector of surrounding points-of-interest (POIs), motivated by studies on region functionality modeling~\cite{Yuan-et-al:KDD2014,Yuan-et-al:TKDE2015}. We compute the functional similarity as: 
\begin{equation}
\alpha_{i, l}^{r_{FS}} = \max (0, Pearson(\mathbf{F}_i, \mathbf{F}_l)).
\label{eq2}
\end{equation}
Here, $ \mathbf{F}_i $ and $ \mathbf{F}_l $ are the POI vectors of observed location $ i $ and unobserved location $ l $, respectively. Each vector records the numbers of surrounding POIs of different POI categories at each location. Function $ Pearson(\cdot, \cdot) $ returns the Pearson correlation coefficient, which is in $ [-1, 1] $, and values smaller than zero are filtered.

(3) \textit{Trans-Probability.}
We further consider the data flow among the locations. For example, people may travel to a distant huge shopping mall center by convenient public transport.
We call the relevance of locations because of the data flow as \emph{Trans-Probability relation}. To showcase the impact of such relations, we consider crowd flows as an example and compute the data transition probability between two locations as:
\begin{equation}
\alpha_{i, l; t}^{r_{TP}} = \frac{\#crowds(v_i \rightarrow v_l|t)}{\#crowds(v_i|t)}.
\label{eq3}
\end{equation}
Here, $ t $ refers to the $ t $-th time step of a day, $ \#crowds(v_i|t) $ denotes the crowd flows at location $ i $ at time step $ t $, and $ \#crowds(v_i \rightarrow v_l|t) $ is the crowd flows transiting from locations $ i $ to $ l $ at time step $ t $. 


\subsubsection{Aggregating information from correlated locations}

Let $ \mathcal{R} $ be the set of aforementioned heterogeneous spatial relations. Next, we aggregate the data from observed locations to estimate those for the unobserved location $ l $ based on each relation $ r \in \mathcal{R} $. To reduce the computation costs and the impact of noisy data, we use the set of the top-$ K $ most correlated observed locations for each unobserved location $ l $ with respect to each relation $ r \in \mathcal{R} $, denoted as $ \mathcal{N}_l^r $. 

We first use fully-connected layers to project the data of the locations in $ \mathcal{N}_l^r $ into a $ D $-dimensional space, denoted as $ \mathbf{H}^r \in \mathbb{R}^{K \times P \times D} $. We aggregate the projected data to generate an inductive hidden representation for an unobserved location $ l $ as:
\begin{equation}
\mathbf{s}_{l,t}^r = \sigma (\sum_{i \in \mathcal{N}_l^r}{\frac{\alpha_{i,l;t}^r}{ \sum_{j \in \mathcal{N}_l^r}{\alpha_{j,l;t}^r} } \mathbf{h}_{i,t}^r \mathbf{W}_s^r + \mathbf{b}_s^r}).
\label{eq4}
\end{equation}
Here, $ \alpha_{i,l;t}^r $ refers to the location relevance uncovered by a type of relation, i.e., $ \alpha_{i, l}^{r_{SP}} $, $ \alpha_{i, l}^{r_{FS}} $, or $ \alpha_{i, l; t}^{r_{TP}} $, where $ \alpha_{i, l}^{r_{SP}} $ and $ \alpha_{i, l}^{r_{FS}} $ are time-invariant, while $ \alpha_{i, l; t}^{r_{TP}} $ is time-dependent; $ \mathbf{h}_{i,t}^r $ is the hidden representation of observed location $ i $ at time step $ t $ with respect to relation $ r $; $ \mathbf{W}_s^r \in \mathbb{R}^{D \times D} $ and $ \mathbf{b}_s^r \in \mathbb{R}^{D} $ are learnable parameters with respect to relation $ r $; $ \sigma(\cdot) $ is a non-linear activation function, e.g., ReLu~\cite{Nair-and-Hinton:ICML2010}. The output of the equation, $ \mathbf{s}_{l,t}^r $, represents the hidden state of an unobserved location $ l $ ($ 1 \leq l \leq M $) at time step $ t $ ($ 1 \leq t \leq P $) with respect to relation $ r $, which aggregates the information from the most correlated locations.

The hidden representation $ \mathbf{s}_{l,t}^r $ in Equation~(\ref{eq4}) mainly considers similarities between the unobserved location $ l $ and each observed location $ i \in \mathcal{N}_l^r $. These locations may not be always highly correlated, and their differences should not be ignored. To this end, we compute the bias between the unobserved location $ l $ and each observed location $ i \in \mathcal{N}_l^r $ at each time step $ t $ as $ |\mathbf{s}_{l,t}^r - \mathbf{h}_{i,t}^r| $. Then, the bias of location $ l $ for relation $ r $ can be formulated as:
%
\begin{equation}
\mathbf{\delta}_{l,t}^r = tanh (\sum_{i \in \mathcal{N}_l^r}{\frac{\alpha_{i,l;t}^r}{ \sum_{j \in \mathcal{N}_l^r}{\alpha_{j,l;t}^r} } |\mathbf{s}_{l,t}^r - \mathbf{h}_{i,t}^r| \mathbf{W}_{\delta}^r + \mathbf{b}_{\delta}^r}),
\label{eq5}
\end{equation}
where $ \mathbf{W}_{\delta}^r $ and $ \mathbf{b}_{\delta}^r $ are learnable parameters. For a weaker correlation, the bias $ \mathbf{\delta}_{l,t}^r $ is larger, and we need to add a greater penalty on $ \mathbf{s}_{l,t}^r $. Thus, the hidden representation of the unobserved location $ l $ at time step $ t $ in relation $ r $ can be adjusted as:
\begin{equation}
\tilde{\mathbf{s}}_{l,t}^r = \sigma (\mathbf{s}_{l,t}^r \mathbf{W}_{\tilde{s}}^r + \mathbf{\delta}_{l,t}^r),
\label{eq6}
\end{equation}
where $ \mathbf{W}_{\tilde{s}}^r $ denotes a learnable parameter matrix. Equation~(\ref{eq6}) captures both the similarities and differences between an unobserved location and observed locations at different time steps, which is beneficial for generating an effective sequence representation for the target unobserved location $ l $ for each relation $ r $. 


%
%

\subsection{Temporal Modeling}

The spatial aggregation stage ignores the temporal correlation of unobserved location time series when forming hidden representation for unobserved location. 
Thus, we further propose a temporal modeling stage to model the temporal correlations in the hidden representations for unobserved locations.

We propose a relation-aware GRU network to exploit related locations' information to guide the temporal dependency modeling process for each relation. The relation-aware GRU network incorporates relational information into the temporal modeling by augmenting GRU with relation-aware gating mechanisms. 


\subsubsection{GRU with context features} 

GRU has shown strong performances in modeling temporal correlations~\cite{Sutskever-et-al:NeurIPS2014}. Since spatio-temporal data is strongly impacted by the context factors such as time, we embed context features (e.g., time of day) into the GRU sequential learning as follows:
\begin{equation}\label{eq7}
\begin{split}
\mathbf{g}_t^r = sigmoid(\mathbf{W}_g^r[\tilde{\mathbf{s}}_{l,t}^r, \mathbf{e}_t, \mathbf{z}_{l, t-1}^r] + \mathbf{b}_g^r), \\
\mathbf{u}_t^r = sigmoid(\mathbf{W}_u^r[\tilde{\mathbf{s}}_{l,t}^r, \mathbf{e}_t, \mathbf{z}_{l, t-1}^r] + \mathbf{b}_u^r), \\
\tilde{\mathbf{z}}_{l, t}^r = tanh(\mathbf{W}_z^r[\mathbf{g}_t^r \otimes \mathbf{z}_{l,t-1}^r, \tilde{\mathbf{s}}_{l,t}^r, \mathbf{e}_t] + \mathbf{b}_z^r), \\
\mathbf{z}_{l,t}^r = (1 - \mathbf{u}_t^r) \otimes \mathbf{z}_{l,t-1}^r + \mathbf{u}_t \otimes \tilde{\mathbf{z}}_{l,t}^r. \\
\end{split}
\end{equation}
Here, $ \tilde{\mathbf{s}}_{l,t}^r $ is the output representation of the spatial aggregation stage for the unobserved location $ l $ at time step $ t $ with respect to relation $ r $; $ \mathbf{e}_t $ denotes the context features at time step $ t $; $ \mathbf{g}_t^r $ and $ \mathbf{u}_t^r $ denote the corresponding reset gate and update gate in the GRU cell, respectively; $ \mathbf{W}_g^r $, $ \mathbf{b}_g^r $, $ \mathbf{W}_u^r $, $ \mathbf{b}_u^r $, $ \mathbf{W}_z^r $, and $ \mathbf{b}_z^r $ are learnable parameters; $ [\cdot,\cdot] $ denotes the concatenation operation; $ \otimes $ denotes the element-wise product operation; and $ \mathbf{z}_{l, t} $ is the GRU output representation. For simplicity, we summarize Equation~(\ref{eq7}) as:
\begin{equation}\label{eq8}
\mathbf{z}_{l,t}^r = \phi (\tilde{\mathbf{s}}_{l,t}^r, \mathbf{e}_t, \mathbf{z}_{l, t-1}^r),
\end{equation}
where $\phi(\cdot)$ denotes the computation function of the GRU cell.

\subsubsection{Relation-aware GRU}

The bias representation $ \delta_{l,t}^r $ is important for guiding the information flow between different time steps at an unobserved location. The bias indicates the difference between the unobserved location and correlated observed locations for each relation, which reflects the reliability of the generated representation at each time step. When the bias grows, the importance of the input representation should decay. Thus, we introduce a relation-aware input gate to adaptively control the importance of the generated representation at each time step:
\begin{equation}\label{eq9}
\beta_{l,t}^r=1/e^{\max(\mathbf{0},\mathbf{W}_{\beta}^r \mathbf{\delta}_{l,t}^r +\mathbf{b}_{\beta}^r)},
\end{equation} 
where $ \mathbf{W}_{\beta}^r $ and $ \mathbf{b}_{\beta}^r $ are learnable parameters. The negative exponential formulation controls the gate $ \beta_{l,t}^r \in (0,1] $. Afterwards, we update the input hidden representation $ \tilde{\mathbf{s}}_{l,t}^r $ by element-wise product with the relation-aware input gate $ \beta_{l,t}^r $.

In addition, the influence of past hidden states should also be adjusted according to the bias to control the information flow between different time steps. Thus, we propose a relation-aware forget gate to adaptively control the influence of past hidden states: 
\begin{equation}\label{eq10}
\gamma_{l,t}^r=1/e^{\max(\mathbf{0},\mathbf{W}_{\gamma}^r \mathbf{\delta}_{l,t-1}^r +\mathbf{b}_{\gamma}^r)},
\end{equation} 
where $ \mathbf{W}_{\gamma}^r $ and $ \mathbf{b}_{\gamma}^r $ are learnable parameters. Then, the GRU hidden state of previous time step $ \mathbf{z}_{l, t-1}^r $ is updated by element-wise product with the relation-aware forget gate $ \gamma_{l,t}^r $.

Finally, the computation of the relation-aware GRU is represented as:
\begin{equation}\label{eq11}
\mathbf{z}_{l,t}^r = \phi (\beta_{l,t}^r \otimes \tilde{\mathbf{s}}_{l,t}^r, \mathbf{e}_t, \gamma_{l,t}^r \otimes \mathbf{z}_{l, t-1}^r),
\end{equation}
where $ \phi(\cdot) $ is the function of standard GRU cell that defined in Equations~(\ref{eq7}) and~(\ref{eq8}). In this way, the relation-aware GRU network is able to adaptively model the temporal correlations for the unobserved location with the guidance of correlated locations' information.

\subsection{Multi-Relation Fusion}

The importance of different relations may differ at different time steps. 
We thus design a multi-relation attention mechanism to dynamically assign different weights to different relations at different time steps. For an unobserved location $ l $, the attention score of relation $ r $ at time step $ t $ is computed as:
\begin{equation}\label{eq12}
a_{l,t}^{r}=\mathbf{v}^\top \tanh(\mathbf{W}_a \mathbf{z}_{l, t}^r + \mathbf{b}_a),
\end{equation}
\begin{equation}\label{eq13}
\lambda_{l,t}^r=\frac{\exp(a_{l,t}^r)}{\sum_{r^{'} \in \mathcal{R}}\exp(a_{l,t}^{r^{'}})},
\end{equation}
where $ \mathbf{z}_{l, t}^r $ is the output representation of the temporal modeling stage, $ \mathbf{W}^a \in \mathbb{R}^{D \times D} $, $ \mathbf{b}^a \in \mathbb{R}^{D} $, and $ \mathbf{v} \in \mathbb{R}^{D} $ are learnable parameters, $ \mathcal{R} $ is the set of heterogeneous spatial relations, and $ \lambda_{l,t}^r $ is the attention score, indicating the importance of relation $ r $ at time step $ t $. Then, the multi-relation information is fused as:
\begin{equation}\label{eq14}
\tilde{\mathbf{y}}_{l,t}=\sum_{r \in \mathcal{R}} \lambda_{l,t}^r \cdot \mathbf{z}_{l, t}^r,
\end{equation}
where $ \tilde{\mathbf{y}}_{l,t} \in \mathbb{R}^{D} $ represents the output representation for the unobserved location $ l $ at time step $ t $, which fuses multi-relation information. Finally, we use two fully-connected layers to compute the kriging output:
\begin{equation}\label{eq15}
\hat{\mathbf{y}}_{l, t} = \mathbf{W}_{y, 2}(\sigma(\mathbf{W}_{y, 1}\tilde{\mathbf{y}}_{l, t} + \mathbf{b}_{y, 1})) + \mathbf{b}_{y, 2},
\end{equation}
where $ \mathbf{W}_{y, 1} \in \mathbb{R}^{D \times D} $, $ \mathbf{b}_{y, 1} \in \mathbb{R}^{D} $, $ \mathbf{W}_{y, 2} \in \mathbb{R}^{D \times C} $, and $ \mathbf{b}_{y, 2} \in \mathbb{R}^{C} $ are learnable parameters, and $ \hat{\mathbf{y}}_{l, t} \in \mathbb{R}^{C} $ denotes the final estimation value for location $ l $ at time step $ t $. 

\subsection{Model Training and Testing}

\begin{figure}
	\centering
	\includegraphics[width=0.9 \columnwidth]{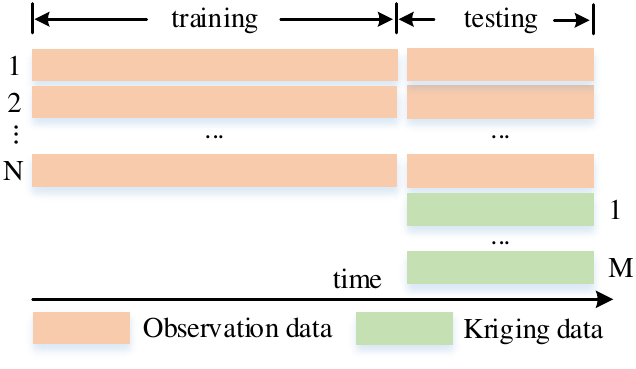} 
	\caption{Data split in model training and testing.}
	\label{fig3}
\end{figure}

We illustrate the model training and testing strategies with Figure~\ref{fig3}. Data is split into training and testing sets in chronological order, and the unobserved locations for testing are unseen at training.

In the training stage, we treat an observed location $ i $ as an unobserved location, and use the time series data of the other $ N - 1 $ observed locations to estimate the values of location $ i $. We repeat this for every observed location and minimize the mean squared error (MSE) between the estimation value $ \hat{\mathbf{x}}_{i, t} $ and the ground truth $ \mathbf{x}_{i, t} $ to optimize our model:
\begin{equation}
\mathcal{L} (\Theta) = \frac{1}{NP}\sum_{i=1}^N\sum_{t=1}^P|| \hat{\mathbf{x}}_{i, t} - \mathbf{x}_{i, t} ||_2^2,
\label{eq16}
\end{equation}
where $ \Theta $ denotes all learnable parameters in INCREASE. Once the model is trained, we can use it to perform kriging for any given unobserved location. Note that the estimation for different locations is independent and can be parallelized for both training and testing.

\begin{table*}
	\centering
	\caption{Kriging performance comparison on the three datasets. SP: spatial-proximity, FS: func-similarity, TP: trans-probability. RMSE, MAE, MAPE: the smaller the better, R2: the greater the better. Numbers in bold denote the best results. Numbers underlined denote the best baseline results. The improvement denotes the performance improvement of INCREASE over the best baseline results.}
	\begin{tabular}{lcccc|cccc|cccc}
		\toprule
		\multirow{2}{*}{Model}	& \multicolumn{4}{c|}{METR-LA}	& \multicolumn{4}{c|}{Beijing}	& \multicolumn{4}{c}{Xiamen}	\\
		& RMSE	& MAE	& MAPE	& R2	& RMSE	& MAE	& MAPE	& R2	& RMSE	& MAE	& MAPE	& R2	\\
		\midrule
		OKriging				& 12.300& 8.383	& 0.236	& 0.681 & 41.076& 25.452& 0.369	& 0.761 & 63.178& 45.814& 0.583 & 0.555 \\
		GLTL					& 9.972	& 6.987 & 0.180	& 0.791	& 37.465& \underline{21.462} & \underline{0.278} & 0.801	& 61.661 & \underline{44.601} & 0.548 & 0.591 \\
		GE-GAN					& 11.248& 7.038	& 0.169	& 0.734	& 41.172& 23.958& 0.313	& 0.760	& 66.614& 46.900& 0.557 & 0.568 \\
		KCN						& 10.817& 7.592 & 0.201 & 0.754 & 40.190& 23.760& 0.310 & 0.771 & 62.414& 45.317& 0.546	& 0.587	\\
		IGNNK					& \underline{9.028}	& \underline{5.917}	& \underline{0.158}	& \underline{0.828} & \underline{36.818} & 21.916 & 0.307 & \underline{0.808} & \underline{61.231} & 44.968 & \underline{0.521} & \underline{0.601}	\\
		\midrule		
		INCREASE (SP) 			& \textbf{8.602} & \textbf{5.425} & \textbf{0.137} & \textbf{0.844}	& 34.552 & 19.584 & 0.247 & 0.831 & 59.321 & 42.530 & 0.497 & 0.613 \\
		INCREASE (SP + FS)  	    & -   	& -  	& -	    & -      & \textbf{34.001} & \textbf{18.851} & \textbf{0.239} & \textbf{0.836} & 58.737 & 42.478 & 0.493 & 0.627	\\
		INCREASE (SP + FS + TP)	& -  	& -		& -    	& -		& -		& -		& -		& -		& \textbf{57.961} & \textbf{41.339} & \textbf{0.481} & \textbf{0.639} \\
		\midrule
		Improvement				& 4.7\% & 8.3\% & 13.3\%& 1.9\% & 7.7\% & 12.2\% & 14.0\% & 3.5\% & 5.3\% & 7.3\% & 7.7\%& 6.3\% \\
		\bottomrule
	\end{tabular}
	\label{table1}
\end{table*}

\section{Experiments} \label{Experiments}


\subsection{Experimental Settings}

\subsubsection{Datasets} 

We conduct experiments on three real-world spatio-temporal datasets. (1) \textbf{METR-LA}~\cite{Li-et-al:ICLR2018} contains traffic speed data collected from 207 sensors in Los Angeles County for four months (March 1, 2012 to June 30, 2012); (2) \textbf{Beijing}~\cite{Zheng-et-al:KDD2015} contains pollutant concentration data (PM2.5) recorded by 36 sensors in Beijing for one year (May 1, 2014 to April 30, 2015); (3) \textbf{Xiamen}~\cite{Zheng-et-al:AAAI2020} contains traffic flow data collected from 95 sensors in Xiamen (a coastal city in China) for five months (August 1, 2015 to December 31, 2015). Appendix~\ref{appendix datasets} summarizes the statistics of the three datasets.

Due to limited data availability, we only use the spatial-proximity relation in the METR-LA dataset. In the Beijing dataset, we use both the spatial-proximity and func-similarity relations. In the Xiamen dataset, we use all three types of relations. The spatial-proximity is computed according to Equation~(\ref{eq1}) using the road network distance for the METR-LA and Xiamen datasets, and Euclidean distance for the Beijing dataset. For the functional similarity, we use surrounding POIs of each category within 300 meters from a location of interest to compute Equation~(\ref{eq2}). To derive the transition probability, we collect the taxis GPS trajectories in Xiamen to compute Equation~(\ref{eq3}). The context features refer to the time feature (time of day) in the experiments.   

Following IGNNK~\cite{Wu-et-al:AAAI2021} (one of the baseline methods), in all datasets, we randomly select 25\% of the sensors as unobserved locations, and the rest are observed locations. We keep the same settings for all methods for fair comparison. We take data from the first 70\% of the time steps for training and test on the rest 30\% of the time steps. We select 20\% of the training set as the validation set for early stopping. Note that the data of unobserved locations are not used in the training or validation process. All data are normalized via the Z-Score method.

\subsubsection{Baselines} 

We compare our proposed model with the following baseline methods: (1) Ordinary kriging (\textbf{OKriging})~\cite{Cressie-and-Wikle:2015} is a well-known spatial interpolation model; (2) Greedy low-rank tensor learning (\textbf{GLTL})~\cite{Bahadori-et-al:NeurIPS2014} is a transductive tensor factorization model for spatio-temporal kriging; (3) \textbf{GE-GAN}~\cite{Xu-et-al:TRC2020} is a transductive method, which combines the graph embedding (GE) technique~\cite{Yan-et-al:TPAMI2007} and a generative adversarial network (GAN)\cite{Goodfellow-et-al:NeurIPS2014}; (4) Kriging convolutional network (\textbf{KCN})~\cite{Appleby-et-al:AAAI2020} is an inductive spatial interpolation method based on graph neural networks; (5) \textbf{IGNNK}~\cite{Wu-et-al:AAAI2021} is an inductive graph neural network for spatio-temporal kriging. We provide detailed description of the baselines in Appendix~\ref{appendix baselines}.

\subsubsection{Evaluation metrics}

We apply four popular metrics to measure the model performance, i.e., root mean squared error (RMSE), mean absolute error (MAE), mean absolute percentage error (MAPE), and R-square (R2), which are defined in Appendix~\ref{appendix evaluation metrics}.

\subsubsection{Implementation details}  

In each dataset, one type of sensor readings is considered, i.e., $ C = 1 $. The numbers of observed locations on the METR-LA, Beijing, and Xiamen datasets are $ N = 157 $, $ N = 27 $, and $ N = 72 $, respectively. The numbers of unobserved locations on the three datasets are $ M = 50 $, $ M = 9 $, and $ M = 23 $, respectively. The length of the time window is set as $ P = 24 $ in all datasets. The sizes of the top neighbor set $ \mathcal{N}_l^r $ for each relation are $ K = 15 $, $ K = 5 $, and $ K = 35 $ for the three datasets, respectively. The dimensionality of the hidden states is set as $ D = 64 $. The nonlinear activation function $ \sigma(\cdot) $ in our model refers to the ReLU activation~\cite{Nair-and-Hinton:ICML2010}. 

We train our model using Adam optimizer~\cite{Kingma-and-Ba:ICLR2015} with an initial learning rate of 0.001 on an NVIDIA GeForce RTX 2080Ti GPU.

\subsection{Experimental Results}

\begin{figure*}
	\centering
	\subfigure[METR-LA]{
		\label{fig4(a)} 
		\includegraphics[width = 0.3 \textwidth]{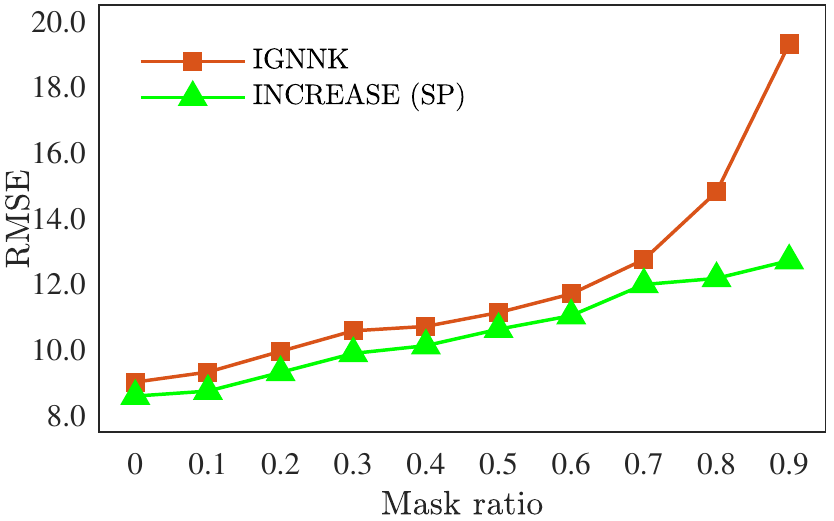}}
	\subfigure[Beijing]{
		\label{fig4(b)} 
		\includegraphics[width = 0.3 \textwidth]{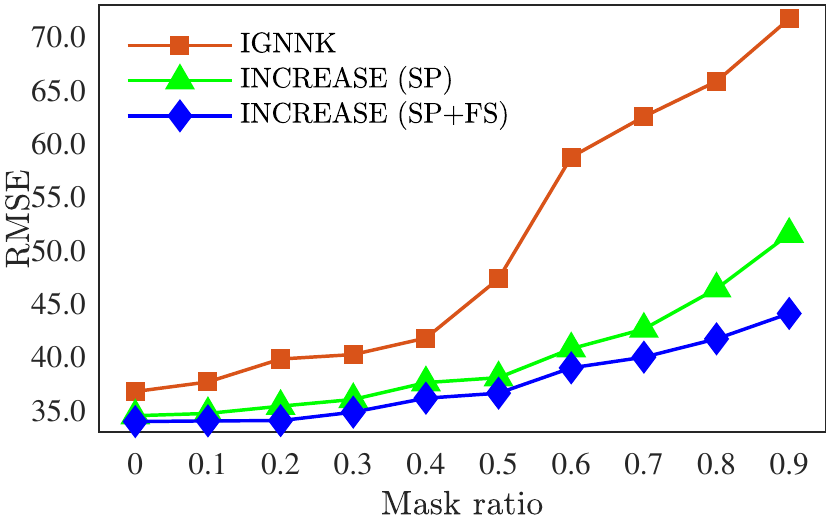}}
	\subfigure[Xiamen]{
		\label{fig4(c)} 
		\includegraphics[width = 0.3 \textwidth]{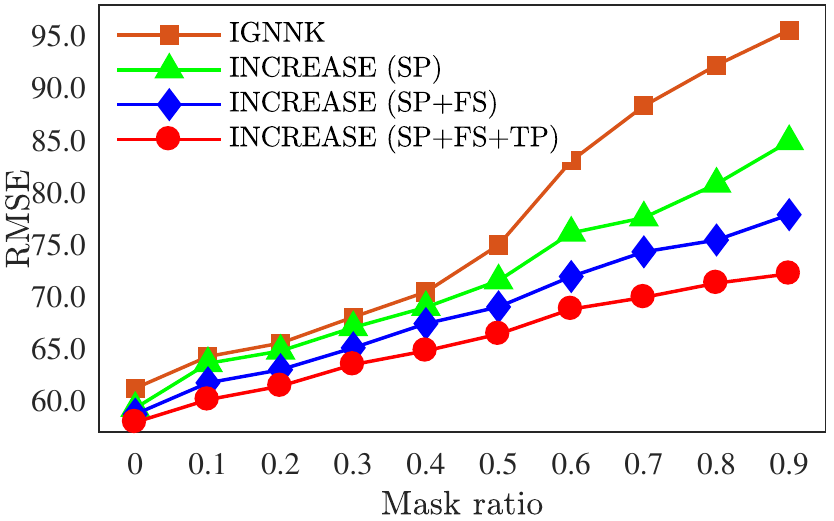}}
	\caption{Model performance when varying the number of observed locations.}
	\label{fig4}
\end{figure*}

\begin{figure*}
	\centering
	\subfigure[Impact of $ K $]{
		\label{fig5(a)} 
		\includegraphics[width = 0.3 \textwidth]{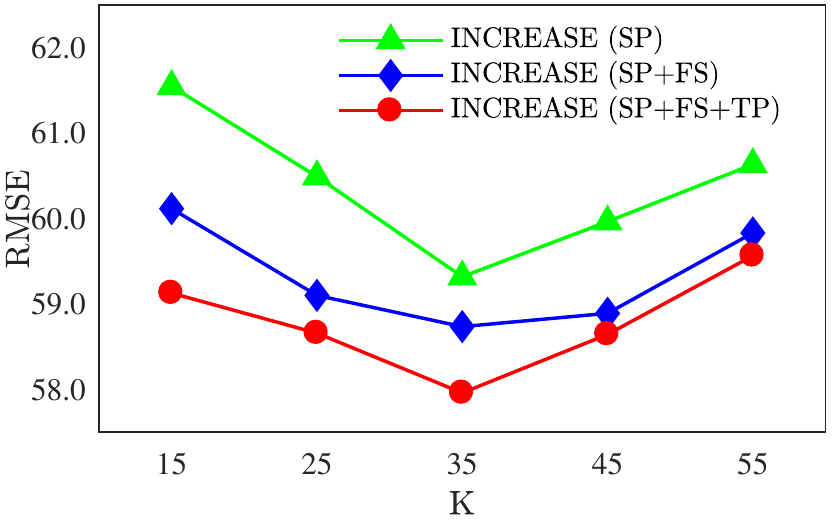}}
	\subfigure[Impact of $ P $]{
		\label{fig5(b)} 
		\includegraphics[width = 0.3 \textwidth]{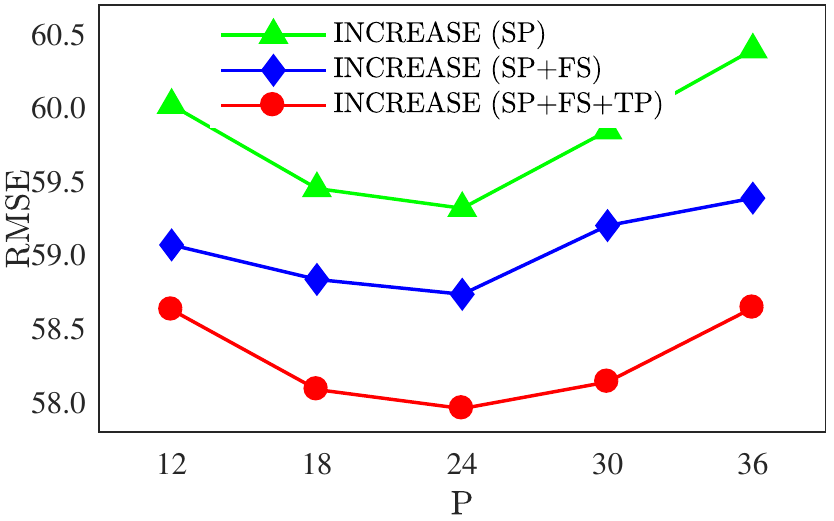}}
	\subfigure[Impact of $ D $]{
		\label{fig5(c)} 
		\includegraphics[width = 0.3 \textwidth]{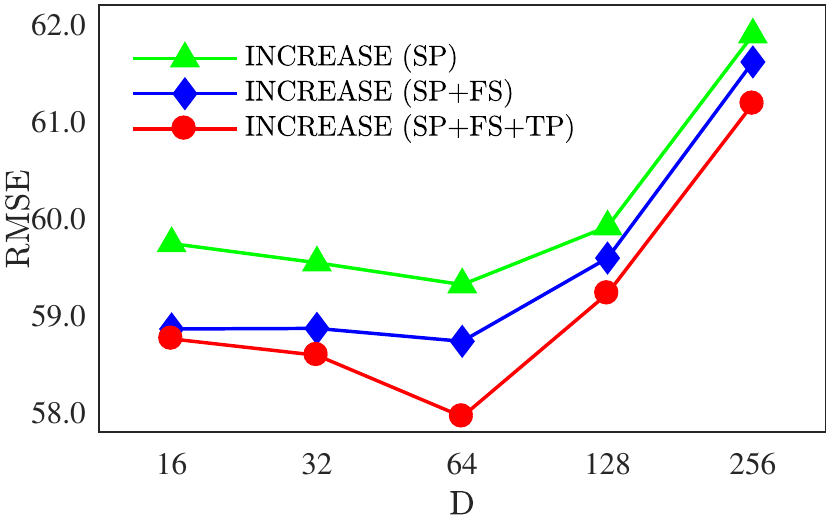}}
	\caption{Parameter study on the Xiamen dataset.}
	\label{fig5}
\end{figure*}

\begin{table}
	\centering
	\caption{Effect of difference modeling (RMSE).}
		\begin{tabular}{lccc}
			\toprule
			Model							& METR-LA			& Beijing			& Xiamen	\\
			\midrule
			w/o difference modeling			& 8.858				& 34.283			& 58.217	\\
			\midrule		
			w/ difference modeling (ours)	& \textbf{8.602}	& \textbf{34.001}	& \textbf{57.961} 	\\
			\bottomrule
		\end{tabular}
	\label{table2}
\end{table}

\subsubsection{Model performance comparison} We first compare the overall performance of INCREASE with baseline methods, and then evaluate the model performance when varying the number of observed locations.

(1) \textit{Overall performance comparison.} Table~\ref{table1} summarizes the overall model performances. Our INCREASE (all three variants) outperforms all baseline methods in terms of all four metrics on the three datasets. OKriging is a traditional kriging method, which suffers in handling the non-linear spatio-temporal data and thus yields poor performances. GLTL is an effective kriging method. However, it is transductive and cannot be directly applied to estimating for new locations of interest without retraining. GE-GAN is also transductive, while its performance is worse than GLTL in general. Indeed, multiple factors contributed to GE-GAN’s worse performance, e.g., it does not distinguish the importance of different observed locations. KCN is an inductive spatial kriging method, which does not consider the temporal correlations. Thus, it also has larger estimation errors than ours. IGNNK uses graph convolutional networks for spatio-temporal kriging, which shows the best performance among the baselines in most cases. However, it only considers the spatial distance relation and fails to model the diverse temporal patterns in the estimation process. Thus, it still performs worse than ours. Overall, INCREASE outperforms the best baseline by up to 14.0\% in terms of the estimation errors, confirming the effectiveness of our model design in capturing the heterogeneous spatial relations and diverse temporal patterns from the data.



(2) \textit{Performance vs. the number of observed locations.} The number of observed locations $ N $ is often quite limited due to the high operating costs of the sensors. To evaluate the impact of $ N $, we randomly drop a fraction $ \eta $ (the \emph{mask ratio}) of observed locations, i.e., using $ N_1 = (1 - \eta) \times N $ observed locations to perform kriging for a fixed number of $ M $ unobserved locations. For fair comparison, we use the same settings of observed locations and unobserved locations for all methods. Figure~\ref{fig4} shows the results. On the METR-LA dataset, we observe that when the mask ratio is larger than 0.7, the performance of IGNNK degrades significantly. IGNNK constructs a graph with $ M + N_1 $ nodes and propagates the information among these nodes. When $ M $ is close to or even greater than $ N_1 $, there is too little reliable information in the graph, and the message passing between the unobserved nodes contributes more errors and thus hinders the model performance. In comparison, our models estimate for each unobserved location independently, which avoids the error propagation problem. On the Beijing dataset, the kriging accuracy of INCREASE (SP + FS) degrades slower than INCREASE (SP), again showing that the functional similarity is helpful for identifying more correlated locations even with limited observed locations. We observe similar results in the Xiamen dataset, demonstrating the advantage of modeling heterogeneous spatial relations for spatio-temporal kriging, especially on more sparse datasets.

\subsubsection{Ablation study} We conduct the following ablation studies to verify the effect of each component in our model.	

(1) \textit{Effect of heterogeneous spatial relations.} In Table~\ref{table1}, INCREASE (SP) denotes using only the spatial-proximity relation. We observe that our model outperforms the baseline models even in this simple variant. This shows the effectiveness of our overall model design. By incorporating the func-similarity relation, INCREASE (SP + FS) improves the performance as it helps the model to find more correlated but distant locations. In the Xiamen dataset, we further model the trans-probability relation, and INCREASE (SP + FS + TP) yields further performance improvements, which confirms the importance of modeling the heterogeneous spatial relations.

\begin{table}
	\centering
	\caption{Effect of relation-aware GRU network (RMSE).}
		\begin{tabular}{lccc}
			\toprule
			Model									& METR-LA			& Beijing			& Xiamen	\\
			\midrule
			w/o $\beta_{l,t}^r$						& 8.741				& 34.400 			& 58.557	\\
			w/o $\gamma_{l,t}^r$					& 8.787				& 34.492			& 58.455	\\
			w/o $\beta_{l,t}^r$ \& $\gamma_{l,t}^r$	& 8.806				& 34.603			& 58.660	\\
			w/o $\mathbf{e}_t$						& 8.692				& 34.251			& 58.150	\\
			\midrule		
			relation-aware GRU (ours)				& \textbf{8.602}	& \textbf{34.001}	& \textbf{57.961} \\
			\bottomrule
		\end{tabular}
	\label{table3}
\end{table}

\begin{table}
	\centering
	\caption{Effect of multi-relation attention (RMSE).}
		\begin{tabular}{lccc}
			\toprule
			Model							& Beijing			& Xiamen		\\
			\midrule
			concat							& 34.314			& 58.498		\\
			add								& 34.621			& 58.525		\\
			\midrule		
			multi-relation attention (ours)	& \textbf{34.001} 	& \textbf{57.961}	\\
			\bottomrule
		\end{tabular}
	\label{table4}
\end{table}

(2) \textit{Effect of difference modeling.} To evaluate the effect of difference modeling in the spatial aggregation stage, we design a variant of INCREASE by removing the difference modeling component (Equations~(\ref{eq5}) and~(\ref{eq6})). The experimental results in Table~\ref{table2} validate that it is beneficial to model the differences among correlated locations for each relation in the spatial aggregation stage. 

(3) \textit{Effect of relation-aware GRU network.} We conduct an ablation study for the relation-aware GRU network in the temporal modeling stage. Table~\ref{table3} shows the experimental results of INCREASE variants with modifications in the relation-aware GRU network. When removing the relation-aware input gate ($\beta_{l,t}^r$) and/or the relation-aware forget gate ($\gamma_{l,t}^r$), the model performance degrades significantly. This demonstrates that the relational information is important for guiding the information flow in the temporal dimension for the unobserved locations. By incorporating the context features ($\mathbf{e}_t$), we observe a consistent performance improvement, confirming their benefits for the kriging.


(4) \textit{Effect of multi-relation attention mechanism.} To evaluate the effect of the multi-relation attention mechanism in the multi-relation fusion stage, we conduct experiments by replacing the multi-relation attention mechanism with simple concatenation or addition approach to fuse the multi-relation information. As only one relation is considered in the METR-LA dataset, we run these experiments on the Beijing and Xiamen datasets. The experimental results in Table~\ref{table4} show the effectiveness of our multi-relation attention mechanism.

\subsubsection{Parameter study}

We study the impact of three parameters: the number of top neighbors $ K $ considered for each relation, the time window length $ P $, and the dimensionality of hidden representations $ D $. Figure~\ref{fig5} presents the experimental results on Xiamen dataset, we provide the results on METR-LA and Beijing datasets in Appendix~\ref{appendix experimental results}.

(1) \textit{Impact of the number of top neighbors $ K $.} As shown in Figure~\ref{fig5(a)}, a larger $ K $ provides more correlated sensors' data for the estimation, which yields better results initially. As $ K $ increases further, farther neighbors are considered, introducing noises and thus negatively impacting the model performance. 

(2)\textit{Impact of the time window length $ P $.} We observe from Figure~\ref{fig5(b)} that the RMSE first decreases and then increases with the increase of the time window length $ P $. This is because a larger $ P $ may offer more training signals which helps the model performance. However, when $ P $ gets too large, it also brings challenges to model the long-term temporal dependencies to make accurate estimations.

(3) \textit{Impact of the dimensionality of hidden representations $ D $.} Figure~\ref{fig5(c)} shows that increasing $ D $ enhances the model’s learning capacity. However, when $ D $ gets larger than 64, the model estimation performance degrades significantly, as the model needs to learn more parameters and may suffer from the over-fitting problem. 

\section{Conclusion} \label{Conclusion}

We proposed INCREASE, an inductive spatio-temporal graph representation learning model for spatio-temporal kriging that can estimate the values for a set of unobserved locations given the data from observed locations. We conducted extensive experiments on three real-world spatio-temporal datasets. Experimental results show that INCREASE outperforms state-of-the-art methods, and the advantage is up to 14.0\% in terms of the estimation errors. The performance gains of INCREASE are more significant when there are fewer observed locations available. For future work, we plan to incorporate more relations (e.g., social interaction among locations) into our model to further improve its performance.

\begin{acks}
The research was supported by Natural Science Foundation of China (62272403, 61872306, 61802325).
\end{acks}

\bibliographystyle{ACM-Reference-Format}
\bibliography{sample-base}

\clearpage

\appendix

\section{Relevance to Web}  \label{appendix relevance to web}

Spatio-temporal graphs have been widely applied in web applications, e.g. social networks~\cite{Zhang-et-al:WWW2022, Schweimer-et-al:WWW2022}, Web of Things applications~\cite{He-et-al:WWW2022,He-and-Shin:WWW2022}. However, in real-world,  
inferring the knowledge for unobserved nodes is extremely challenging due to the heterogeneous spatial relations and diverse temporal patterns, which largely limits the web and social graph applications of existing methods~\cite{Appleby-et-al:AAAI2020,Wu-et-al:AAAI2021}. 
Thus, the inductive graph representation learning model for spatio-temporal kriging (INCREASE) studied in this paper, aiming to estimate the data for unobserved locations using the observation data, essentially addresses a core challenge of the web -- improving the spatio-temporal kriging performance with inductive graph representation learning, to enable the web as a technical infrastructure for web and social applications. It will also enhance the understanding of proactiveness and inclusiveness of social web analysis and graph algorithms.

\begin{table}
	\centering
	\caption{Table of notations.}
	\resizebox{\columnwidth}{!}{
		\begin{tabular}{c|l}
			\toprule
			Symbol						& Definition	\\
			\midrule
			$ P $						& the time window length	\\
			$ N $						& the number of observed locations	\\
			$ M $						& the number of unobserved locations	\\
			$ i $						& an observed location	\\
			$ l $						& an unobserved location	\\
			$ r $						& a type of relation	\\
			$ \alpha_{i,l;t}^r $		& \begin{tabular}[l]{@{}l@{}}the location relevance between locations $ i $ and $ l $ at time\\ step $ t $ for relation $ r $\end{tabular}	\\
			$ \beta_{l,t}^r $			& the relation-aware input gate	\\
			$ \gamma_{l,t}^r $			& the relation-aware forget gate \\
			$ \lambda_{l,t}^r $			& the attention score of location $ l $ at time step $ t $ for relation $ r $	\\
			$ \mathbf{x}_{i,t} $		& the observation data of location $ i $ at time step $ t $	\\
			$ \mathbf{e}_{t} $			& the context features at time step $ t $	\\
			$ \mathbf{h}_{i,t}^r $		& \begin{tabular}[l]{@{}l@{}}the hidden representation of location $ i $ at time step $ t $ for\\ relation $ r $\end{tabular}	\\
			$ \mathbf{\delta}_{l,t}^r $	& the bias of location $ l $ at time step $ t $ for relation $ r $	\\
			$ \tilde{\mathbf{s}}_{l,t}^r $	& the output representation of the spatial aggregation stage	\\
			$ \mathbf{z}_{l,t}^r $		& the output representation of the temporal modeling stage	\\
			$ \tilde{\mathbf{y}}_{l,t} $& the output representation of the multi-relation fusion stage \\
			$ \hat{\mathbf{y}}_{l, t} $	& the estimated value for location $ l $ at time step $ t $	\\ 
			\bottomrule
	\end{tabular}}
	\label{table5}
\end{table}

\section{Notations} \label{appendix notations}

We present some important notations used in this paper in Table~\ref{table5}.



\section{Experimental Setup}

\subsection{Datasets} \label{appendix datasets}

\begin{table*}
	\centering
	\caption{Summary statistics of the three datasets.}
	\begin{tabular}{lcccccc}
		\toprule
		Dataset	& Time period	& Time interval	& \# Time steps	& \# Sensors & Measurement 			\\
		\midrule
		METR-LA	& 1 March 2012 - 30 June 2012	& 5-minute	& 34,272	& 207	& Traffic speed		\\
		Beijing	& 1 May 2014 - 30 April 2015	& 1-hour	& 8,760		& 36	& PM2.5				\\
		Xiamen	& 1 August 2015 - 31 December 2015	& 5-minute	& 44,064	& 95	& Traffic flow		\\
		\bottomrule
	\end{tabular}
	\label{table6}
\end{table*}

\begin{figure*}
	\centering
	\subfigure[METR-LA]{
		\label{fig6(a)} 
		\includegraphics[width = 0.64 \columnwidth]{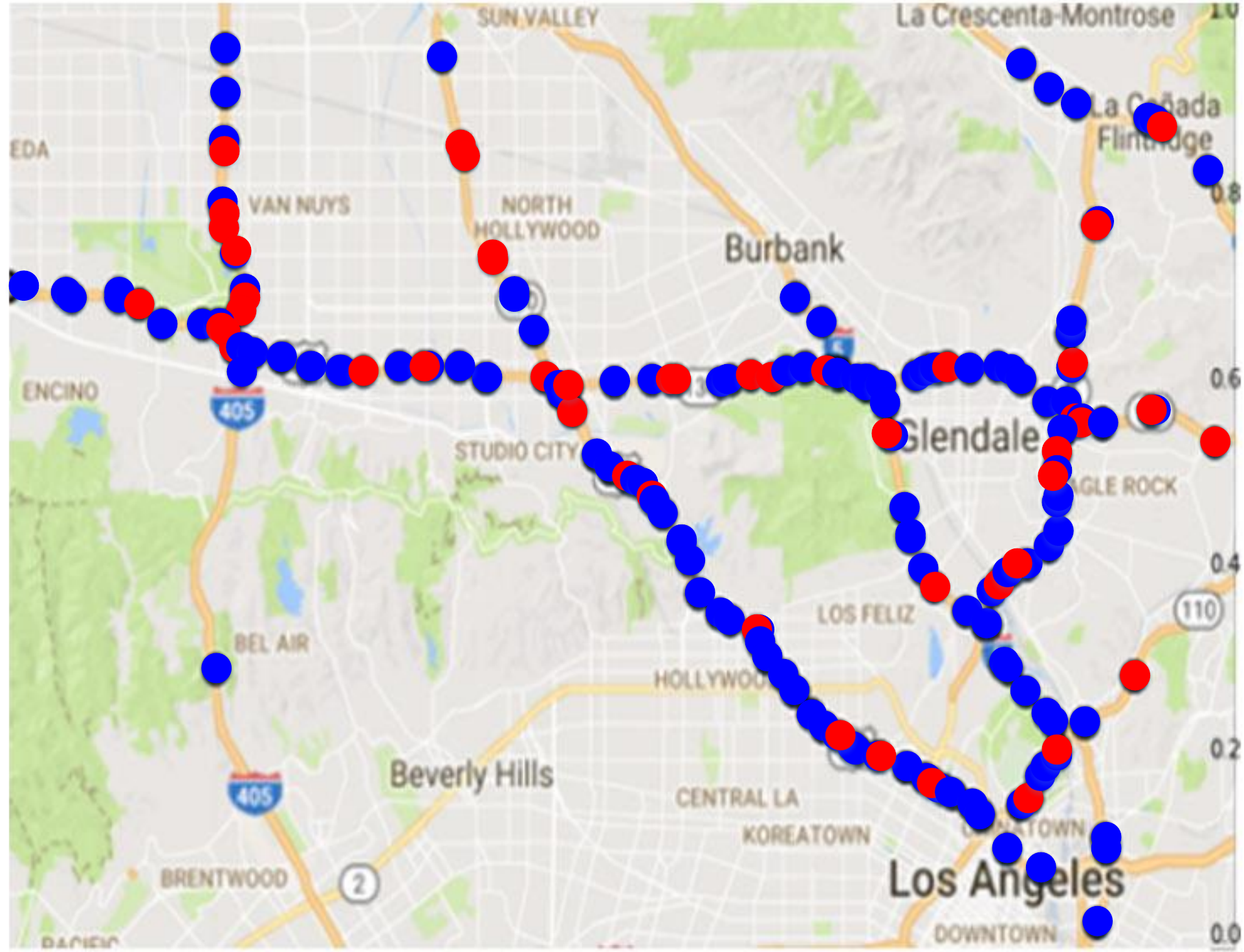}}
	\subfigure[Beijing]{
		\label{fig6(b)} 
		\includegraphics[width = 0.64 \columnwidth]{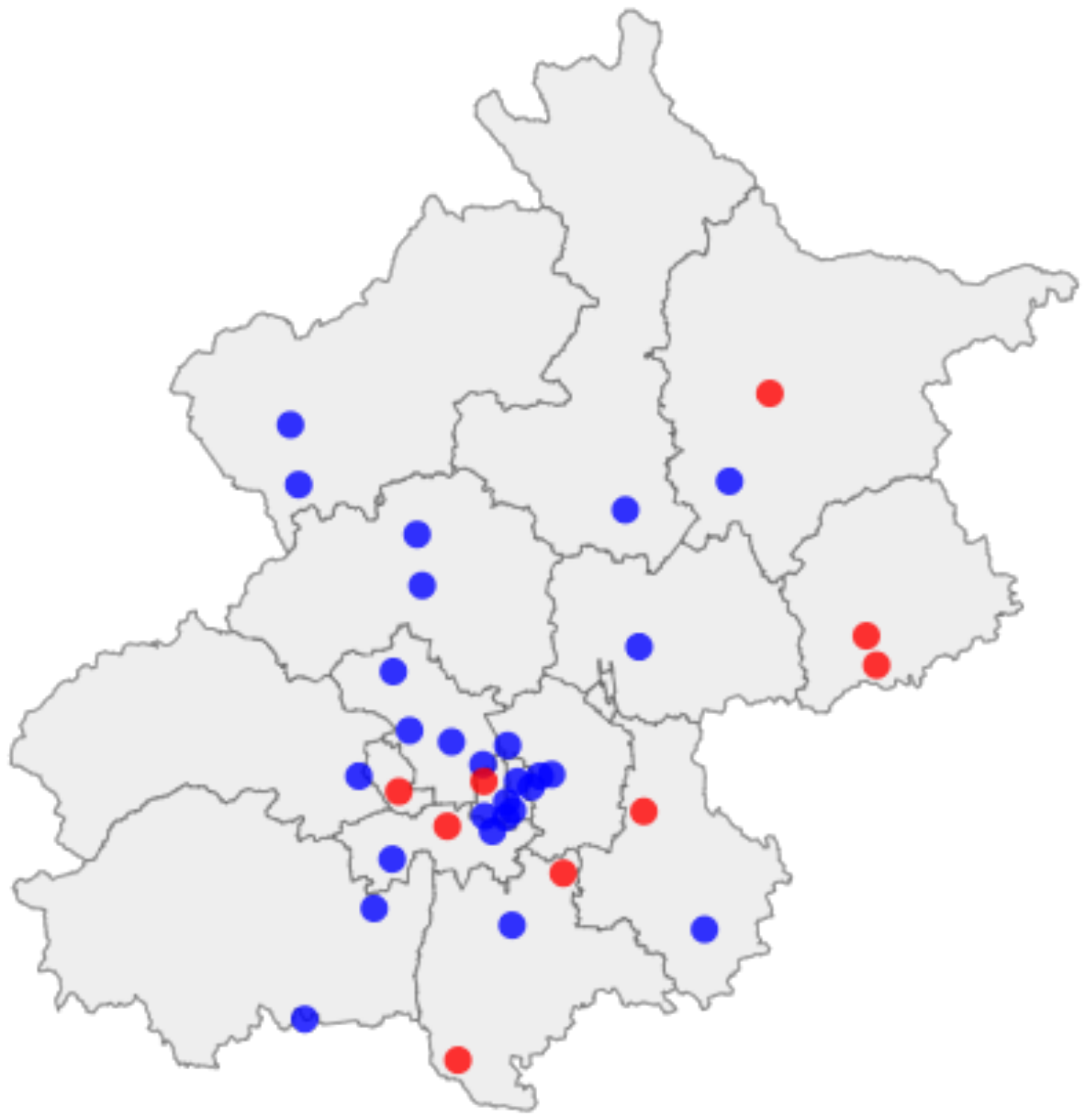}}
	\subfigure[Xiamen]{
		\label{fig6(c)} 
		\includegraphics[width = 0.64 \columnwidth]{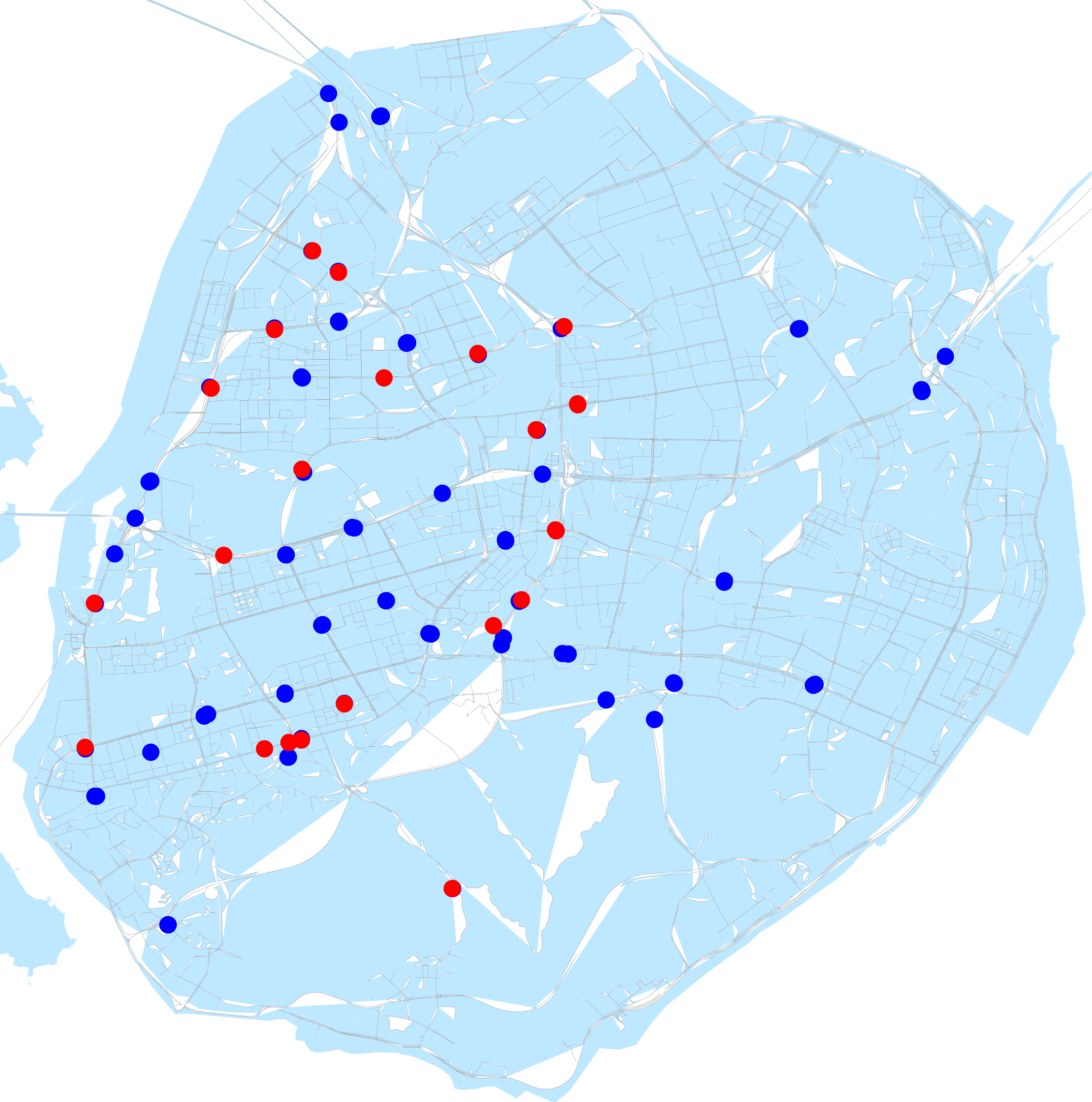}}
	\subfigure{ 
		\includegraphics[width = 1.0 \columnwidth]{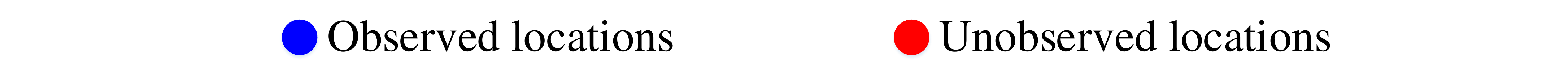}}
	\caption{Distribution of observed locations and unobserved locations of the three datasets.}
	\label{fig6}
\end{figure*}

We present statistics of the three datasets in Table~\ref{table6}, and visualize the distribution of observed locations and unobserved locations of the three datasets in Figure~\ref{fig6}. 

\subsection{Baselines} \label{appendix baselines}

We compare our proposed model INCREASE with the following baseline methods.

\begin{itemize}
	\item Ordinary kriging (\textbf{OKriging})~\cite{Cressie-and-Wikle:2015} is a well-known spatial interpolation model. We test a number of variograms (i.e., linear, power, gaussian, spherical, exponential, hole-effect) and choose the best based on the smallest residual sum of squares.
	\item Greedy low-rank tensor learning (\textbf{GLTL})~\cite{Bahadori-et-al:NeurIPS2014} is a transductive tensor factorization model for spatio-temporal kriging. GLTL takes an input tensor of shape $ location \times time \times variables $ with zeros at the unobserved locations, and it recovers the values for the observed locations via tensor completion.
	\item \textbf{GE-GAN}~\cite{Xu-et-al:TRC2020} is a transductive method, which combines the graph embedding (GE) technique~\cite{Yan-et-al:TPAMI2007} and a generative adversarial network (GAN)\cite{Goodfellow-et-al:NeurIPS2014}. It uses GE to select the most correlated locations and GAN to generate estimations for the unobserved locations. 
	\item Kriging convolutional network (\textbf{KCN})~\cite{Appleby-et-al:AAAI2020} is an inductive method based on graph neural networks. It is essentially a spatial interpolation method, i.e., it performs kriging for each time step independently. There are three variants of KCN~\cite{Appleby-et-al:AAAI2020}: (1) KCN based on graph convolutional networks~\cite{Kipf-and-Welling:ICLR2017}, (2) KCN based on graph attention networks~\cite{Velickovic-et-al:ICLR2018}, and (3) KCN based on GraphSAGE~\cite{Hamilton-et-al:NeurIPS2017}. We use the best one for testing based on the validation loss. 
	\item \textbf{IGNNK}~\cite{Wu-et-al:AAAI2021} is an inductive graph neural network for spatio-temporal kriging. It constructs a spatial graph with both observed and unobserved locations as the nodes, in which the values at different time steps are considered as features of the nodes (0's for the unobserved locations). Then, it conducts graph convolutions for message passing to recover the values for the unobserved locations.
\end{itemize}

\subsection{Evaluation Metrics} \label{appendix evaluation metrics}

We apply four widely used metrics to measure the model performance, i.e., root mean squared error (RMSE), mean absolute error (MAE), mean absolute percentage error (MAPE), and R-square (R2), which are defined as follows.

\begin{equation}
	RMSE = \sqrt{\frac{1}{MP}\sum_{l=1}^{M}\sum_{t=1}^{P}(\hat{\mathbf{y}}_{l,t}-\mathbf{y}_{l,t})^2},
	\label{eq17}
\end{equation}
\begin{equation}
	MAE = \frac{1}{MP}\sum_{l=1}^{M}\sum_{t=1}^{P}|\hat{\mathbf{y}}_{l,t}-\mathbf{y}_{l,t}|,
	\label{eq18}
\end{equation}
\begin{equation}
	MAPE = \frac{1}{MP}\sum_{l=1}^{M}\sum_{t=1}^{P}|\frac{\hat{\mathbf{y}}_{l,t}-\mathbf{y}_{l,t}}{\mathbf{y}_{l,t}}|,
	\label{eq19}
\end{equation}
\begin{equation}
	R2 = 1 - \frac{\sum_{l=1}^{M}\sum_{t=1}^{P}(\hat{\mathbf{y}}_{l,t}-\mathbf{y}_{l,t})^2}{\sum_{l=1}^{M}\sum_{t=1}^{P}(\bar{\mathbf{y}}-\mathbf{y}_{l,t})^2},
	\label{eq20}
\end{equation}
where $ \hat{\mathbf{y}}_{l,t} $ and $ \mathbf{y}_{l,t} $ are the estimated value and ground truth of location $ l $ at time step $ t $, respectively, and $ \bar{\mathbf{y}} = \frac{1}{MP}\sum_{l=1}^{M}\sum_{t=1}^{P} \mathbf{y}_{l,t} $ in Equation~(\ref{eq20}) is the average value of the ground truth.

\section{Parameter Study on the METR-LA and Beijing Datasets} \label{appendix experimental results}

\begin{figure*}
	\centering
	\subfigure[Impact of $ K $]{
		\label{fig7(a)} 
		\includegraphics[width = 0.32 \textwidth]{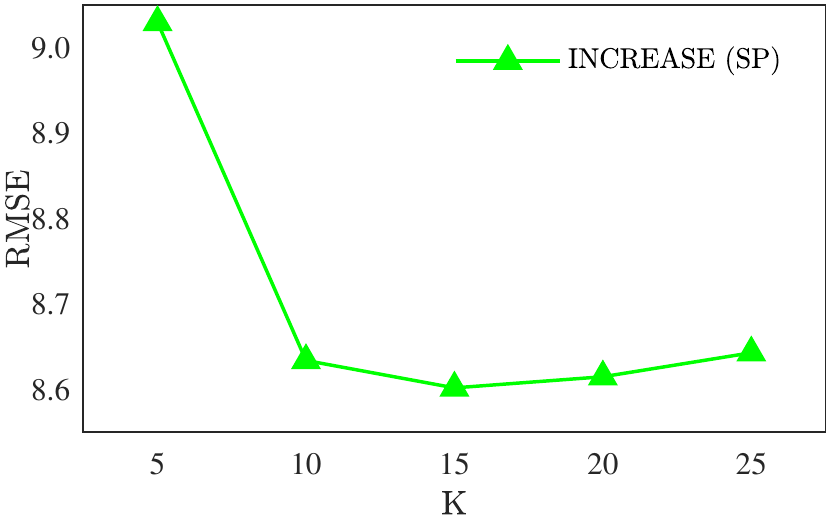}}
	\subfigure[Impact of $ P $]{
		\label{fig7(b)} 
		\includegraphics[width = 0.32 \textwidth]{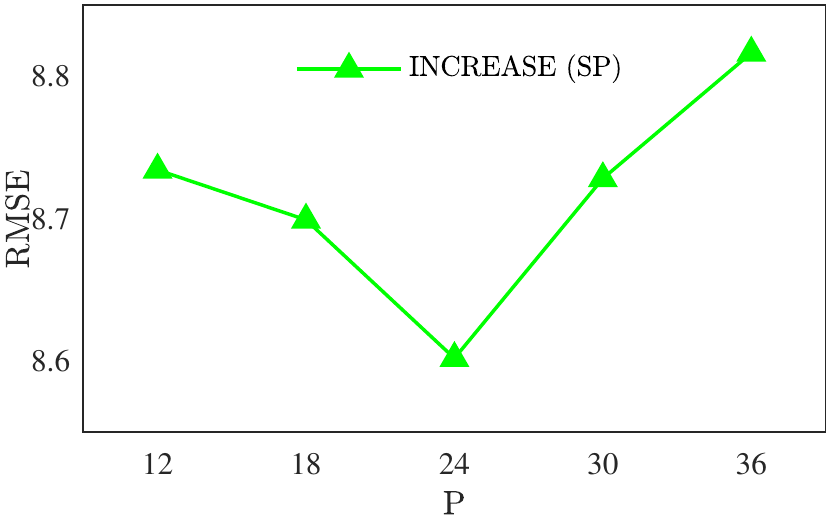}}
	\subfigure[Impact of $ D $]{
		\label{fig7(c)} 
		\includegraphics[width = 0.32 \textwidth]{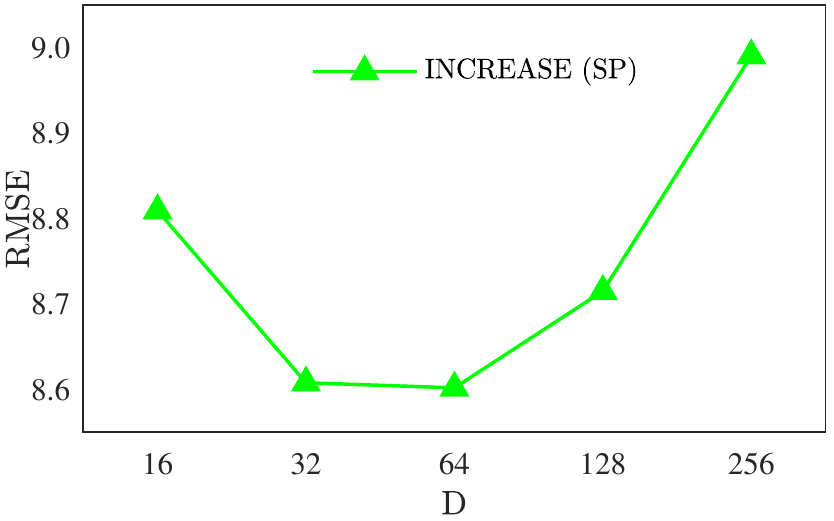}}
	\caption{Parameter study on the METR-LA dataset.}
	\label{fig7}
\end{figure*}

\begin{figure*}
	\centering
	\subfigure[Impact of $ K $]{
		\label{fig8(a)} 
		\includegraphics[width = 0.32 \textwidth]{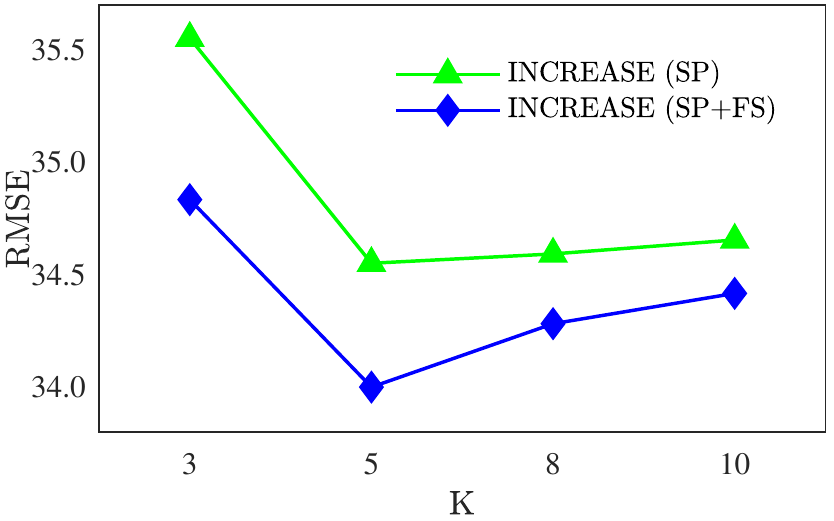}}
	\subfigure[Impact of $ P $]{
		\label{fig8(b)} 
		\includegraphics[width = 0.32 \textwidth]{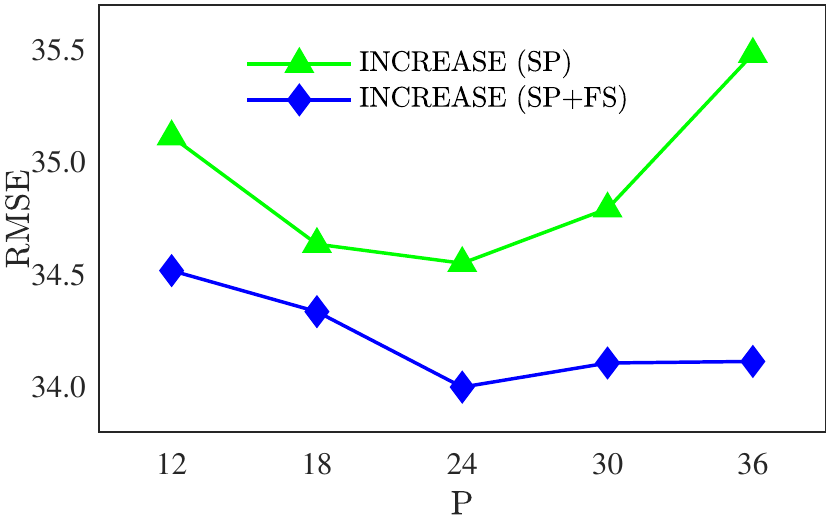}}
	\subfigure[Impact of $ D $]{
		\label{fig8(c)} 
		\includegraphics[width = 0.32 \textwidth]{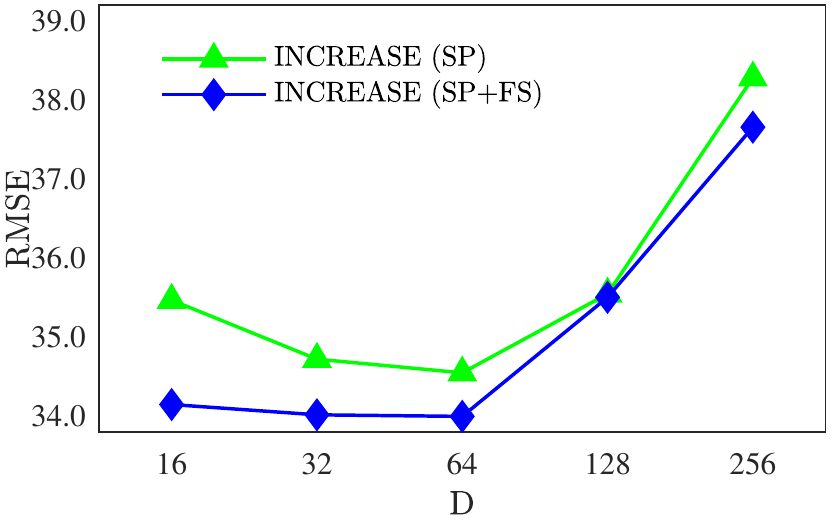}}
	\caption{Parameter study on the Beijing dataset.}
	\label{fig8}
\end{figure*}

We present the parameter studies on METR-LA and Beijing datasets in Figures~\ref{fig7} and~\ref{fig8}, respectively.

\end{document}